\documentclass[11pt,a4paper]{article}
\usepackage[hyperref]{acl2021}
\usepackage{times}
\usepackage{latexsym}

\usepackage{microtype}

\usepackage{booktabs}
\usepackage{graphicx}

\aclfinalcopy 

\title{Tapping BERT for Preposition Sense Disambiguation}

\author{
 Siddhesh Pawar$^*$ \qquad
 Shyam Thombre$^*$ \qquad
 Anirudh Mittal\thanks{~~The three authors contributed equally to the paper} \\
  CFILT, IIT Bombay, Mumbai \\
  \texttt{siddheshpawar1999@gmail.com},
  \texttt{shyam.thombre@iitb.ac.in}, \\
  \texttt{anirudhmittal@cse.iitb.ac.in}\\ 
 \AND
 Girishkumar Ponkiya \qquad
 Pushpak Bhattacharyya \\
  CFILT, IIT Bombay, Mumbai \\
  \texttt{\{girishp,pb\}@cse.iitb.ac.in} \\}

\date{}

\begin{document}
\maketitle

\begin{abstract}
 Prepositions are frequently occurring polysemous words. Disambiguation of prepositions is crucial in tasks like semantic role labelling, question answering, text entailment, and noun compound paraphrasing. In this paper, we propose a novel methodology for preposition sense disambiguation (PSD), which does not use any linguistic tools.  In a supervised setting, the machine learning model is presented with sentences wherein prepositions have been annotated with `senses'. These `senses' are IDs in what is called `The Preposition Project (TPP)'. We use the hidden layer representations from pre-trained BERT and BERT variants. The latent representations are then classified into the correct sense ID using a Multi Layer Perceptron. The dataset used for this task is from SemEval-2007 Task-6. Our methodology gives an accuracy of 86.85\% which is better than the state-of-the-art.
\end{abstract}


\section{Introduction} \label{sec:intro}
Prepositions are among the foremost commonly used terms and they the most ambiguous words in English \cite{baldwin2009prepositions}. According to the British National Corpus \cite[BNC]{bnc-corpus}, prepositions account for four of the top 10 most frequently used terms in English (\textit{of}, \textit{to}, \textit{in}, and \textit{for}).They can impart different context to other parts of the sentences i.e. noun, verbs etc. It can be tricky to identify what’s the meaning or `sense' of the preposition. 

The Preposition Project (TPP) \cite{DBLP:journals/corr/abs-2104-08922} provides a comprehensive characterization of English preposition senses suitable for use in natural language processing. Each of 673 preposition senses for 334 prepositions (mostly phrasal prepositions) has been described by giving it a semantic role or relation name and by characterizing the syntactic and semantic properties of its complement and attachment point. Each sense is further described by its definition and sample usages from the Oxford Dictionary of English, its position in a semantic hierarchy of prepositions, its basic syntactic placement (as described in A Comprehensive Grammar of the English Language), other synonymic prepositions filling a similar semantic role, FrameNet frames and frame elements used to describe the complement, other syntactic forms in which the semantic role may be realized. 

This ambiguity renders the task of disambiguation of prepositions significant. It helps in semantic role labeling \cite{ye2006semantic}, where the task is to identify predicates, extract their arguments, and label the arguments with predefined semantic roles. In most cases, the predicate is a verb, and argument is a noun phrase (subject) or a preposition phrase (direct object or indirect object). Consider the following example:
\begin{quote}
    \textit{John ate some rice \textbf{with$_1$} dal \textbf{with$_2$} a spoon \textbf{with$_3$} his friend.}
\end{quote}

\begin{table}[!ht]
\resizebox{\columnwidth}{!}{%
 \begin{tabular}{ccll} \toprule
  \textbf{\#} 
  & \textbf{Sense ID} 
  & \textbf{Relation} 
  & \textbf{Complement} \\ \midrule
  
  1 & 1(1) & Accompanier 
  & \begin{tabular}[l]{@{}l@{}}anything that can accompany\\ the attachment point\end{tabular} \\ \hline
  
  2 & 4(3) & Means
  & \begin{tabular}[c]{@{}l@{}}an instrument in \\ the action described by \\ the attachment point\end{tabular} \\ \hline
  
  3 & 9(7) & Concomitant 
  & \begin{tabular}[c]{@{}l@{}}sb or sth  linked with \\ subject via the POA\end{tabular} \\ \bottomrule
\end{tabular}}
\caption{Senses of the three occurrences of \textit{with} in the above example. (sb:somebody; sth: something)}
\label{tab:senses-of-with}
\end{table}

Table \ref{tab:senses-of-with} explains how a prepositions can take different meanings. TPP gives several insights for each sense for detailed analysis. Select terms like Sense ID, semantic relation and complement properties have been noted for the above example. Sense IDs are essentially labels defined by TPP. The integer outside the bracket refers to the super sense. The integer inside the bracket refers to the sense of the preposition. Mapping for supersenses and list of all possible meanings for any preposition can be found on the TPP website\footnote{\url{https://www.clres.com/}}.

Understanding the sense of the preposition can help us understand fine semantic relations. This can further be instrumental in other NLP tasks like question-answering. In the example, once senses of  \textit{with} (Accompanier, Means, Concomitant) is known, answers to questions like - \textit{What does John eat with rice?} (dal), \textit{What does John eat rice with?} (spoon), \textit{Who does John eat with?} (friend) become straightforward.
Further, SemEval-2013 Task4 \cite{hendrickx2013semeval} discusses that prepositions are a preferred choice when it comes to an understanding and expressing a relation between the components of a noun compound. Following up on this, \newcite{ponkiya2018treat} states that prepositional paraphrasing is a crucial step in the paraphrasing of noun compounds. Similarly, the application of PSD can also be seen in text entailment, phrasal verb paraphrasing, and so on.

\begin{figure*}[t!]
    \centering
    \includegraphics[width=0.8\textwidth]{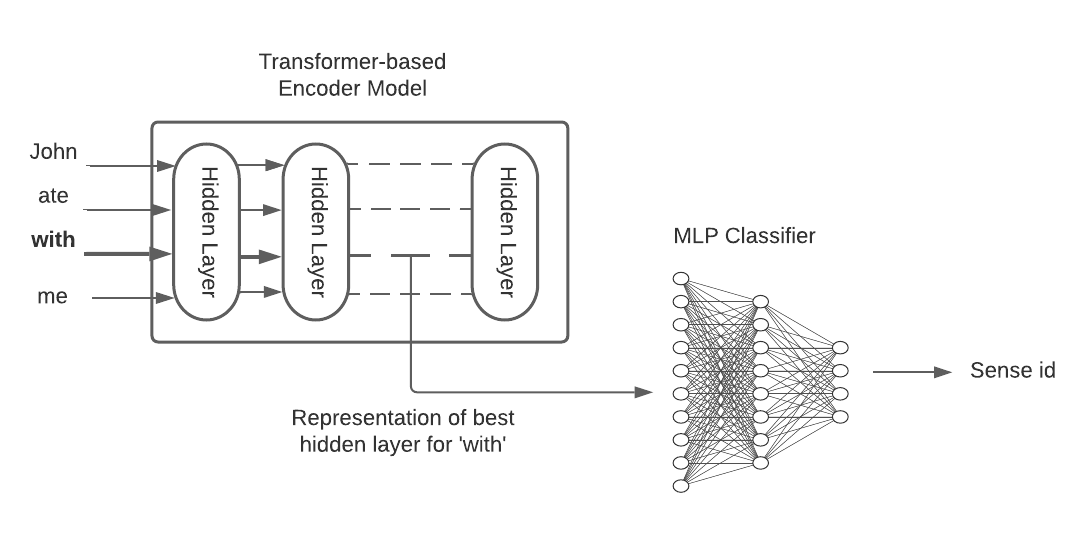}
    \caption{Model architecture for preposition sense disambiguation. The figure shows hidden layers of the transformer-based encoder model, and the arrows shows the possible latent representations. One of the hidden layers is chosen (for each preposition) for representation during testing, marked by the bold arrow}
    \label{fig:architecture}
\end{figure*}

Early works on Preposition sense disambiguation (PSD) like \newcite{hovy2011models}, \newcite{ye2007melb} involved linguistic tools and resources like dependency parser and/or POS taggers. The current state-of-the-art by \newcite{litkowski} also leveraged lemmatizer, dependency parser, along with WordNet. \newcite{gong2018preposition} introduced a new concept free from the use of linguistic resources (except word representations). It made use of Word2Vec \cite{mikolov2013efficient} which used left and right context to generate embedding. However, using a static word representation made the use of hyper-parameter tuning necessary. 

In this paper, we make use of more recent word representations like BERT \cite{devlin2019bert} and its derivatives like RoBERTa \cite{liu2019roberta} and Big Bird \cite{zaheer2020big}. The work by \cite{du2019using} suggests using different layers of language models learn things differently. We identified the respective hidden layer for each preposition that maximizes the output and used that layer as input to the classifier. Our approach achieved slightly better accuracy than the current state-of-the-art while without using any linguistic tool.

The rest of the paper is organized as follows: Section 2 discusses different methods tried for PSD starting from classical machine learning algorithms and heavy linguistic resources to more recent developments. In section 3, we detail the approach used in this paper. Section 4 explains the dataset and other details of the experiment. In section 5, we analyze how different language models perform. This also includes examination of misclassified samples. 
We discuss results and analysis in Section 6, followed by a conclusion and potentially fruitful direction.
\section{Related Work} \label{sec:related-work}

Preposition sense disambiguation has been significantly explored using various classic machine learning techniques. They have used heavy linguistic tools and resources such as part of speech taggers, chunkers, dependency parsers, named entity extractors, Word-Net based super-sense taggers, and semantic role labelers for feature engineering \cite{yuret2007ku,tratz2011fast,srikumar2013modeling}.

\newcite{litkowski} used lemmatizer, dependency parser as well as Word-net to extract features to get the senses of the prepositions to get state-of-the-art accuracy of 85.7\% on the task of preposition sense disambiguation using the SemEval-2007 dataset.
\newcite{gonen2016semi} used signals from the un-annotated multilingual parallel corpus using sequence to sequence neural networks for preposition disambiguation and achieved accuracy within 5\% of the state-of-the-art, which has been achieved in \newcite{litkowski}, \newcite{srikumar2013modeling} and \newcite{hovy2010s}.
Recently, \newcite{gong2018preposition} used combined word vectors of left and right context as well as a context interplay vector as features to train the sense classifier. The context interplay vector  is the vector closest to the sub-spaces spanned by the left and right context word vectors.  \newcite{hassani2017disambiguating} used deep convolutional neural networks along with lexical and syntactic features as well as word embeddings for sense disambiguation of spatial prepositions.

BERT \cite{devlin2018bert} is a language representation model that has been designed to pre-train contextualized deep representations of words based on large unlabelled corpora. 
The BERT model has been proven to be effective in extracting word features and contextual information from plain text. The contextualized BERT embeddings have been
shown to be capable of clustering polysemic words into distinct sense regions in the embedding space \cite{wiedemann2019does}.
Among the work done on the use of embeddings for word sense disambiguation (WSD) is \newcite{peters2018deep} that incorporated the pre-trained ELMo embeddings as WSD features. A study by \newcite{du2019using} fine-tuned BERT and used various internal representations from the BERT encoder as features for WSD. 
In one more study, \newcite{zhu2020cross} used pre-trained embeddings from mBERT \cite{pires2019multilingual} along with a dependency parser to train a classifier for WSD in a cross-lingual setting in SemEval-2021 Task 2. \newcite{huang2019glossbert} incorporated sense definitions (gloss) into supervised WSD to define a new framework called GlossBERT.

\begin{table*}[!ht]
\resizebox{\textwidth}{!}{%
 \begin{tabular}{llc} \toprule
 \textbf{System}
 & \textbf{Techniques Used}
 & \textbf{Accuracy(\%)} \\\hline
 
 \newcite{ye2007melb} 
 & chunker, dependency parser, named entity extractor, WordNet
 & 69.3 \\
 
 \newcite{litkowski} 
 & lemmatizer, dependency parser, WordNet 
 & 85.7 \\
 
 \newcite{gonen2016semi} 
 & multilingual corpus, aligner, dependency parser
 & 81.3 \\
 
 \newcite{gong2018preposition}
 & Word2Vec with fixed context size
 & 80.0 \\ \hline
 
 Our System
 & Pretrained Transformers
 & \textbf{86.9} \\ \bottomrule
 \end{tabular}}
 \caption{\label{tab:res2}Performance of our system for PSD compared with reported results.}
\end{table*}

\section{Our Approach} \label{sec:approach}

The crux of our approach lies in using the embeddings learnt by the pre-trained transformer models. We work on the hypothesis that the latent representations learnt by pre-trained transformer models have `sense discriminative' capabilities, for each preposition. So, fundamentally we have an `encoder model' that obtains the meaningful latent representations or embeddings, and a classifier attempts to classify the embedding of the preposition into one of its senses. Thus we train a separate classifier for each of the prepositions with the obtained embeddings as the input data. The architecture of each classifier is the same, which is a Multi-Layer Perception network. The visualization of our approach can be seen in Figure \ref{fig:architecture}.

\subsection{Encoder model}
We use BERT and its variant to get a contextual representation of a preposition. We provide a sentence as input to the model, and get representation of the preposition from different layers. We use a development set to decide which hidden-layer representation is more appropriate for each of the prepositions.

We are specifically interested in the hidden state output corresponding to the preposition token. Let preposition $p$ is at $i$-th token in the input sentence $S$. We get a representation $v_{ij}$ from $j$-th layer.
\begin{equation}
    v_{ij} = BERT(S, i, j)
\end{equation}
We treat $j$ (hidden layer number) as a hyper-parameter and use a development set to fine-tune it for each preposition.

\subsection{MLP Classifier}
We feed the contextual representation $v_{ij}$ to a multi-layer perceptron (MLP) to predict sense of the preposition. The MLP uses two hidden layer to compute the following:
\begin{equation}
    h = ReLU(FF_1(v_{ij}))
\end{equation}
\begin{equation}
    out = softmax(FF_2(h))
\end{equation}
where, $FF_1$ and $FF_2$ are two feed-forward layers. The softmax function after the final layer ensures that we get the probability distribution over senses. We predict a sense with the maximum probability as the correct preposition.

\subsection{Loss and Optimizer}
We train the MLP classifier with the cross-entropy loss, which is given by:
\begin{equation}
    L= -\frac{1}{N}\sum_{n=1}^{N}\sum_{s=1}^{S_k}\textbf{y}_n^s \log \hat{\textbf{y}}_n^s
\end{equation}
where, $\textbf{y}_n$ is the one-hot encoding of the true sense of $n$-th example, and $\hat{\textbf{y}_n}$ is the predicted probability distribution for the $n$-th example. Minimizing the cross entropy loss is equivalent to maximizing the log likelihood of the training data, thus improving the performance on the test data.

For the optimization, we make use of the \textit{Adam} optimizer with $\beta_1$ as 0.9 and $\beta_2$ equal to 0.999 (default parameters). The loss convergence was observed to the stable when using the \textit{adam} optimizer, which also performed well even for the test data.

\section{Experiments} \label{sec:experiments}
 In our approach to PSD, we require the sentence and the preposition as input to the model. We have used the pre-trained transformer models to generate the latent representation of the preposition in the sentence. We experiment with BERT-base, BERT-large, DistilBERT, Big Bird, RoBERTa, and ALBERT. The experiments consisted of choosing the best hidden layer of transformers for extracting the latent representation. 

\subsection{Dataset}
We use the SemEval-2007 Task 6 dataset for testing our methodology for PSD. The corpus consists of 24,633 sentences in total. There are 34 prepositions in the dataset, and the total numbers of sense are equal to 332. There is no data for 75 senses, and the number of senses per preposition varies. Preposition \textbf{on} has highest number (25) of senses while the preposition \textbf{as} has the lowest number (2) of senses.

\subsection{Training and Evaluation}
Since we have used the pre-trained transformer models for extracting representations for the prepositions, we froze the transformers' parameters during the MLP classifier training. Hence, the training only included optimising the parameters of the Multi-Layer Perceptron classifier for all the senses of each preposition.  
For evaluating the overall performance on the task, we find the accuracy of individual classifiers. We then compute the average over all prepositions for comparison with the baseline.
\section{Results and Discussions} \label{sec:results}
The results for our experiments using various variants of the BERT model are shown  in the table \ref{tab:res}. The classifier for each representation was trained with the hidden layer that gave the best accuracy. The best accuracy did not necessarily come from the last layer. The classifier based on the representations from Big Bird  performs the best amongst all the variants. The comparison of our system with other systems has been shown in table \ref{tab:res2}, and as can be seen, our system outperforms all the existing systems. Additionally, our system is language-agnostic and does not require linguistic tools or static word embedding, unlike other work done in PSD.

\begin{table}[!ht]
\centering
\begin{tabular}{lr}
\toprule
{\bf Model}   & {\bf Accuracy(\%)} \\ \midrule
BERT - base   & 85.4 \\ \hline
BERT - large  & 86.1 \\ \hline
DistilBERT    & 81.5 \\ \hline
RoBERTa       & 83.8 \\ \hline
Big Bird       & 86.9 \\ \hline
ALBERT        & 83.4 \\ \hline
\end{tabular}
\caption{\label{tab:res}Performance of our system with representations from various different pre-trained models.}
\end{table}

Some prepositions like \textit{through}, \textit{on} and \textit{after} give accuracy less than 75\%. We studied the misclassified samples to understand why the model is unable to learn their senses well. One big reason is the data imbalance. This can be validated while comparing the prepositions \textit{on} and \textit{of}. They both have a similar number of classes (20 and 17 respectively), but \textit{of} has almost 4 times the data than \textit{on}. Given this, it can be predicted that \textit{of} should have significantly higher accuracy than \textit{on}. Surprisingly, accuracy for \textit{of} (0.84) is faintly better than accuracy for \textit{on} (0.83). After deeper analysis, it appears that though \textit{of} has the highest number of datapoints per sense in the dataset, the distribution is highly skewed. 9 out of 17 senses of \textit{of} get only 8\% of the data, while the 4 most frequent sense enjoy about 75\% of the data. Many sense with few datapoints (\~ 10) were mostly classified as a 3(1b), a sense with high number (\~ 700) of data. For even preposition \textit{on}, there are cases of improper distribution, but much fewer than \textit{of} which creates the disparity in the result. 

However, it was observed in several cases that if the difference of data between two senses was not large, the model failed to distinguish between the senses properly. For example, among the senses of preposition \textit{of}, the sense 11(6) represents \textit{`noun representing the subject of action denoted in the POA'} and the sense 12(6a) represents \textit{`noun representing the object of action denoted in the POA'}. The former sense was many times predicted as the latter suggesting that the model fails to identify the finer nuance in their meanings.

On the contrary, if the meaning of the sense is starkly different from others, it can be easy for the model to correctly predict the sense. For example, for the preposition \textit{above}, the sense $9(3)$ only had 8 sentences, but still got 100\% accuracy during testing. The reason behind this becomes clear after looking at how all the 5 senses of \textit{above} are defined. Sense 9(3) complements an \textit{established norm} or a \textit{specified amount} like `\textit{above average}', `\textit{above \$ 19}', '\textit{above 90\%}'. This makes it easier for the model to identify the pattern and differentiate among senses.

Data augmentation is one of the ways widely used in deep learning systems to increase the amount of data available. The sense of the preposition depends on the properties of the point of attachment and the complement of the preposition. Thus, as long as the attachment and compliment properties remain the same, the sense does not change. For every sentence in the training data, we can replace one or more words in the point of attachment and complement (preferably nouns) in the sentence with other words with the same properties. 
This phenomenon has been shown in \ref{tab:res3} wherein even after substituting the nouns in the sentence, the sense of preposition \textit{of} remains the same. In these examples, the point of attachment is a location, and the complement is also a location, so when we replace the location with a thing (in this case \textit{football}), the sense changes. 
This property gives an important pointer to data augmentation, especially for the senses with very little or no data, thereby balancing the dataset. 
We are currently working on automating the process of identifying a set of words that retain the properties of point of attachment and complement and plan to include it in future work.

\begin{table}[t]
 \centering
 \begin{tabular}{lc} \toprule
  \textbf{Sentence} & \textbf{Sense} \\ \midrule
  Mumbai is the capital of India      & 6(3) \\
  Chicago is the capital of India     & 6(3) \\ \hline
  India is the capital of Football    & 9(5) \\ 
  India is the capital of Mumbai      & 6(3) \\ \bottomrule
 \end{tabular}
 \caption{\label{tab:res3}Effect of substitution on the senses}
\end{table}

%

\section{Conclusion and Future Work} \label{sec:conclusion}
  This paper proposes a transformer-based method for the task of preposition sense disambiguation, which only relies on pre-trained language models. Different variants of the BERT model were tried, and Big Bird gave the best results on the task due to the long length of sentences in the train as well as the test set.
 The proposed method gives state-of-the-art results on the standard preposition sense disambiguation task without relying on any linguistic machinery, thus drastically reducing the human effort required for the task. This methodology can also be extended to low resource languages where there is an absence of linguistic resources as the BERT model is trained on the un-annotated text corpus.
 
 In the future, we would like to investigate data augmentation techniques to expand the training dataset. The substitution of point of attachment or complement with `similar' words seems promising. Another important future direction can be incorporating gloss pairs that are sense definitions in the BERT model to check whether GlossBERT can effectively deal with the high level of polysemy that is present in the case of prepositions. 

\bibliographystyle{acl_natbib}
\bibliography{main}

\end{document}